\documentclass[journal,12pt,onecolumn,draftclsnofoot,]{IEEEtran}
\usepackage{bm,cite,algorithm,algorithmic,float,amsmath,amssymb}

\usepackage{amssymb}
\usepackage{amsmath}
\usepackage{graphicx}
\usepackage{cite}
\usepackage{citesort}
\usepackage{url}
\usepackage{amsthm}
\usepackage{framed}

\usepackage{siunitx}
\IEEEoverridecommandlockouts

\usepackage{graphicx,epstopdf}
\usepackage{epsfig}
\usepackage{amsfonts}
\floatname{algorithm}{Algorithm}

\newcounter{TodoCounter}

\newcounter{QstCounter}

\newcounter{meNoteCounter}

\newcounter{jdNoteCounter}

\newcounter{jmNoteCounter}

\newcounter{mjNoteCounter}

\begin{document}

\title{Liftago On-Demand Transport Dataset and Market Formation Algorithm Based on Machine Learning}

\author{\IEEEauthorblockN{Jan Mrkos\IEEEauthorrefmark{1}, Jan Drchal\IEEEauthorrefmark{1}, Malcolm Egan\IEEEauthorrefmark{2} and Michal Jakob\IEEEauthorrefmark{1}}

\IEEEauthorblockA{\IEEEauthorrefmark{1}\footnotesize{Department of Computer Science, FEE, CTU in Prague, Czech Republic\\
Email: jan.mrkos$\vert$jan.drchal$\vert$jakob@agents.felk.cvut.cz}}

\IEEEauthorblockA{\IEEEauthorrefmark{2}\footnotesize{Laboratoire de Math{\'e}matiques, UMR 6620 CNRS, Universit{\'e} Blaise Pascal, France\\
Email: malcolm.egan@gmail.com}}
}

\maketitle

\begin{abstract}
This document serves as a technical report for the analysis of on-demand transport dataset. Moreover we show how the dataset can be used to develop a market formation algorithm based on machine learning. Data used in this work comes from Liftago, a Prague based company which connects taxi drivers and customers through a smartphone app. The dataset is analysed from the machine-learning perspective: we give an overview of features available as well as results of feature ranking. Later we propose the SImple Data-driven MArket Formation (SIDMAF) algorithm which aims to improve a relevance while connecting customers with relevant drivers. We compare the heuristics currently used by Liftago with SIDMAF using two key performance indicators.
\end{abstract}

\section{The Liftago Dataset}
Data used in this report were provided by Liftago company. It consists of transactions which were recorded using the smartphone application as well as server-side generated. The dataset is a sample of collected data recorded during 2015 in Prague, Czech Republic.

\begin{figure}[!ht]
\centering
\includegraphics[width=0.8\textwidth]{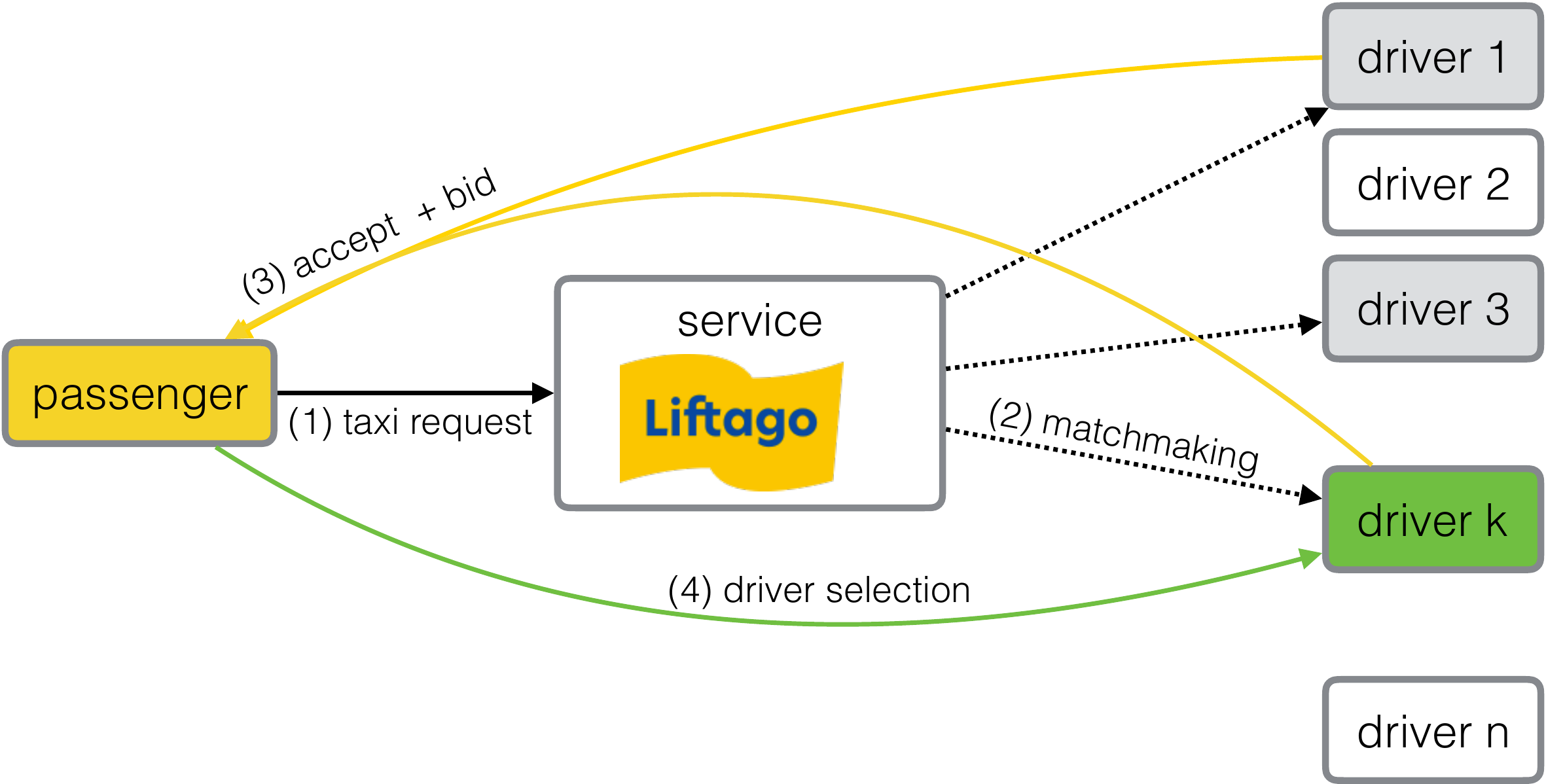}
\caption{Liftago service overview. It takes four steps to issue a ride order: passenger sends a request (1), Liftago designates a set of drivers (2), subset of which accepts the request and offers a price (3), finally the passenger selects a single offer. See text for more information.}
\label{fig:liftago_service}
\end{figure}

Liftago market formation hybrid mechanism is depicted in Figure~\ref{fig:liftago_service}. A potential passenger initiates the process of matchmaking by issuing a ride order using the smartphone application (1). The trip is described by a pickup and possibly a drop off location. Liftago in turn designates a set of drivers which are addressed using a driver-side mobile application (2). Each driver can either accept or reject the request or the request can time out. Accepting drivers provide their price offers (bids) back to the passenger (3). In the last step the passenger can accept one of the offers and finally select the driver.

In the dataset, each transaction is initiated by a passenger request and the following data are recorded: time of the request, pickup location, drop off location (optional), ids of drivers which were offered the journey, initial driver locations and driver responses. Additionally the dataset is supplemented with GPS locations of available drivers sampled every 20 seconds.

In order to develop our data-driven market formation algorithm, we extracted a number of features from the Liftago transactions dataset. The overview of the features is given in Table~\ref{table:features}. All features were extracted directly from the transactions with an exception of \verb"mean_accept_rate" which was derived from per-driver aggregations. We used N\ang{50;5.284;} E\ang{14;25.246;} as the coordinates of Prague centre. The dataset used in the following section is based on \num{31787} ride orders which corresponds to \num{253687} instances (requests), where the number of requests accepted by the drivers is \num{84384} while the number of declined and timed out requests is \num{169303}. The ride orders were exchanged between 5647 active customers (customers that finished a ride) and 390 active drivers. Liftago released a smaller version of the dataset for free\footnote{\url{http://try.liftago.com/info-wants-to-be-free-en/}}.


\begin{table}
\centering
\caption{Features derived from the Liftago transactions dataset.}
\begin{tabular}{|c|c|}
  \hline
  \textbf{Feature} & \textbf{Description} \\
  \hline
  \verb"pickup_distance" & direct Euclidean distance to pickup (km) \\
  \hline
  \verb"ride_distance" & direct Euclidian distance from pickup to destination (km) \\
  \hline
  \verb"pickup_center" & pickup Euclidian distance from the Prague center (km) \\
  \hline
  \verb"ride_center" & pickup Euclidian distance from the Prague center (km) \\
  \hline
  \verb"hour" & time of day (h) \\
  \hline
  \verb"day" & day of the week (0 - 6) \\
  \hline
  \verb"mean_accept_rate" & driver's mean accept rate over all transaction records \\
  \hline
\end{tabular}\label{table:features}
\end{table}

\section{Acceptance Model}\label{sec:model}

After identifying key features in the Liftago transaction dataset, we developed an ACceptance Model (ACM). This binary classification model provides a means of predicting the probability that a driver will accept (and successively bid) for a passenger's request based on features shown in Table~\ref{table:features}. The two classes: driver's accept and driver's reject/timeout are well balanced in the dataset with accepted requests making up \(52\%\) of all requests in the dataset.

The classifier was based on Random Decision Forest Ensemble (RDFE)~\cite{Breiman2001} as implemented in~\cite{scikit-learn}. RDFE gave best results when compared to other machine learning approaches. Moreover, it serves as a feature ranking method where the feature importances are derived by averaging expected fraction of samples affected by tree nodes over all trees. The forest consisted of 200 trees, Gini impurity criterion was employed. The model's accuracy and F1 score were estimated to \num{0.781} and \num{0.775} using 5-fold cross-validation. For further use in Section~\ref{sec:sidmaf} it was trained using all available data.

\begin{figure}[!ht]
\centering
\includegraphics[width=0.6\textwidth]{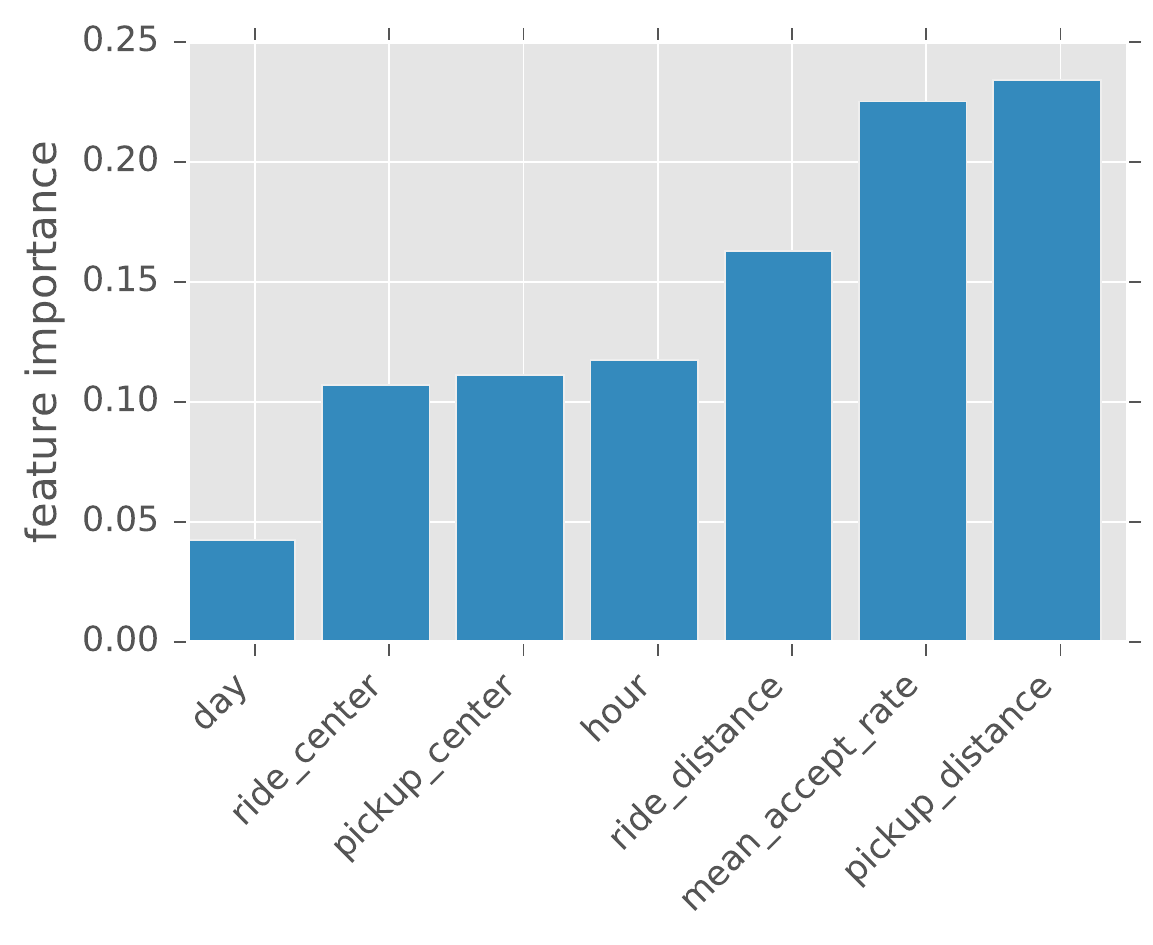}
\caption{Feature ranking of Liftago data. The bars show ACM model feature importances obtained using RDFE based on 200 trees. The most important features are those approximating ride order distances (\texttt{pickup\_distance} and \texttt{ride\_distance}) and \texttt{mean\_accept\_rate} based on per-driver aggregated historical data.}
\label{fig:ranking}
\end{figure}

The results of feature ranking are shown in Figure~\ref{fig:ranking}. A key observation is that the distance-based \verb"pickup_distance" and \verb"ride_distance" features are among the top three features, which is consistent with intuition. Notably, the driver history in terms of the aggregated \linebreak\texttt{mean\_accept\_rate} is the second highest ranked feature.



Apart from the predicted classes, RDFE provides class probabilities for evaluated samples. We use these probabilities in the following section, where the ACM becomes a vital part of our market formation algorithm.

\section{Data-Driven Market Formation Algorithm}
\label{sec:sidmaf}


In this section we will show how the Acceptance Model (ACM) developed in the previous section can be used in a simple data driven market formation algorithm called SIDMAF (SImple Data-driven MArket Formation).

The outline of the algorithm is simple: its goal is to select a smallest subset of currently available drivers such that the probability of at least one driver accepting (and hence bidding) is greater than a preselected threshold $p_T$. The motivation for the algorithm is straightforward: 1) addressing the smallest possible subset of drivers implies highly relevant recommendation and less distraction for drivers, while 2) setting $p_T$ gives a probabilistic guarantee for the service availability (and hence passenger satisfaction).

SIDMAF operation per each passenger ride order can be described in the five following steps:
\begin{enumerate}
\item Apply the trained ACM from Section~\ref{sec:model} to compute the probability $p_i$ that driver $i$ will bid for each of $n$ currently available drivers.
\item Sort the probabilities in decreasing order; i.e., $p_1 \geq p_2 \geq \cdots \geq p_n$.
\item A probability of at least a single driver accepting out of $l$ first drivers is given by $p^l_{1+,acc} = 1 - p^l_{rej}$, where $p^l_{rej}$ is a probability of none of the first $l$ drivers accepting. More precisely $p^l_{rej} = \prod_{i=1}^{l}(1 - p_i)$.
\item Find the smallest $l$ such that $p^l_{1+,acc} > p_T$. Set $l=n$ if $p_{1+,acc}^n \leq p_T$.
\item Offer the passenger request to the ranked drivers in $\{1,2,\ldots,l\}$.
\end{enumerate}

With a simple modification to the third step, SIDMAF can select a smallest subset of drivers such that the probability of accept by \(k\) or more drivers is higher than the predetermined threshold \(p_T\). Probability of accept by \(k\) or more drivers can be calculated as:
\begin{equation}
p^l_{k+,acc} =  p^l_{1+,acc} - \sum_{i=1}^{k-1}p^l_{i,acc},
\end{equation}
where \(p^l_{k+,acc}\) is the probability of \(k\) or more accepts and \(p^l_{i,acc}\) is the probability of exactly \(i\) accepts in \(l\) drivers. Because Liftago's heuristics aims to present each customer with at least three offers, it is reasonable for SIDMAF to prioritize three accepts while minimizing the number of approached drivers for the comparison.

The SIDMAF can be evaluated in two possible ways: Firstly, it can be deployed in  the real-world environment and an approach like A/B testing can be used to assess its strengths and weaknesses. Secondly, its evaluation can be based on a simulation. Although the first option should be the preferred one from the accuracy point of view, it is also the riskier and more expensive one.

In this paper we employ the later simulation-based approach using the already available data: we use the recorded ride orders to generate demand as well as positions of all available drivers to work with a realistic spatio-temporal distribution. This allows not only a reasonable estimation of SIDMAF performance but also a seamless comparison to Liftago's current market formation approach. The details of Liftago's market formation algorithms are not publicly available but unlike our data-driven method they are significantly dependent on an expert knowledge.

The current heuristics used by Liftago aims to provide each customer with an offer while simulatneously keeping the number of approached drivers low. Sometimes, the heuristic may result in spamming drivers, where some drivers will receive large number of requests in a short period of time due to their relative position to multiple customers. The main motivation for SIDMAF is to use historical data to select smallest possible subsets of relevant drivers, which would reduce spam, while providing probabilistic guarantees for the number of offers received by the customers.

In the course of simulation we go through all requests $R$ in the dataset. For a request $r \in R$, SIDMAF selects a set $T_r$ of $k_r$ drivers from a set of all currently available drivers. This selection is defined by the threshold \(p_T\). After we have selected the drivers, we randomly choose a single driver (which corresponds to an offer) $t_r \in T_r$ as accepted by the customer. This driver is in turn removed from the set of available drivers for the duration of the ride. In this simulation, we do not model drivers' bids. The selected driver is considered to have accepted the request and customer has selected his offer. This is irregardless of the drivers modelled accept probability.

For the purpose of the simulation, where our goal is to show that SIDMAF can reduce the spam to the drivers while providing a guaranteed number of offers to each customer, the random selection is a sufficient approximation of the driver-customer negotiation. To model this interaction fully, we would have to first model which drivers accepted the request. For each accepting driver, we would have to predict his bid. Finally, based on the offer and the parameters of the ride order, customer model should be used to predict which offer the customer chooses. While our experiments with customer models were promising, we do not have sufficient data to model the bids made by the drivers.

There are many ride orders not completed in the original dataset.  This can be because the customer is only probing the market with no intention to actually order a taxi, because the customer does not receive any favourable offer or because the customer receives no offer at all. Effectively, we do not have a good way to distinguish between the first two causes of incomplete ride orders. Only \(43\%\) of ride orders are concluded in a ride. This means that if we were to use all ride orders for the simulation, we would soon have more than twice the number of ride orders being serviced by the available taxis. This would lead to an unrealistic saturation of available drivers. For this reason we simulate only ride orders which actually concluded in the ride.

The locations of the drivers used in the simulation are the real-world locations as recorded in the dataset. When a selected driver is removed from the set of available drivers for the duration of the ride, his position at the end of the ride will be his position in the real-world dataset, not the final destination of the simulated ride\footnote{This is true only in the case where the randomly selected driver is not the same as the driver selected by the customer in the dataset.}

To calculate the durations of the rides, we used an average speed of all taxi rides recorded in the dataset multiplied by an euclidean distance between the rides origin and destination.

To compare the heuristic used by Liftago and SIDMAF, we define two Key Performance Indicators (KPIs) in order to evaluate the market formation approaches. These KPIs quantitatively describe efficiency of the matchmaking as well as the average level of spam to the drivers:
\begin{enumerate}
\item $KPI1$: Average ratio of accepts per ride order:
\begin{align}
KPI1 = \frac{1}{|R|k_r}\sum_{r\in R}\sum_{i=1}^{k_r} p_{i}^{r},
\end{align}
where $p_{i}^{r}$ is the accept probability for the $i$-th driver for the ride order $r$. To evaluate $KPI1$ directly on Liftago dataset we set $p_{i}^{r} = 1$ for accepting and $p_{i}^{r} = 0$ for non-accepting driver. Higher value of $KPI1$ means more successful matchmaking.
\item $KPI2$: Average number of selected drivers per ride order:
\begin{align}
KPI2 = \frac{1}{|R|}\sum_{r\in R} k_r,
\end{align}
where smaller value means less spam for drivers.
\end{enumerate}

Tables~\ref{table:results_1} and \ref{table:results_2} show the performance of the SIDMAF algorithm compared to the real-world Liftago transactions. Observe that our market formation algorithm performs better than the existing approach adopted by Liftago in both cases.
In Table \ref{table:results_1}, we present the results of running SIDMAF in its simpliest form, that is, when selecting a set of drivers so that the probability of at least one driver accepting is above the threshold \(p^1_T\), in this case \(p^1_T=0.999\).
Table \ref{table:results_2} gives results of running SIDMAF in its extended form, when selecting a set of drivers so that the probability of at least 3 drivers accepting is above the threshold \(p^3_T = 0.9\).


As can be observed from the results, the simulation shows that the performance of SIDMAF is better than the heuristic currently used by Liftago. Notably, even though the extended SIDMAF selects almost twice as many drivers as the basic SIDMAF on average ($KPI2$), $KPI1$ differs only slightly between the two versions of the algorithm, in both cases outperforming the Liftago heuristic by a large margin. The number of selected drivers ($KPI2$) is lower for both SIDMAF version where for the basic version the difference is significant. 

The promising results indicate that it might be beneficial for Liftago to perform a real-world evaluation of SIDMAF in order to obtain more precise estimates of its properties.

\begin{table}
\centering
\caption{Comparison of real-world transactions by Liftago to SIDMAF with \(p_T^1=0.999\), probability of at least one driver accept}
\begin{tabular}{l|cc}

   Market formation & KPI1 & KPI2 \\
  \hline
  Liftago & 0.476  & 7.67 \\
  SIDMAF & \textbf{0.867} & \textbf{3.56}\\

\end{tabular}\label{table:results_1}
\end{table}

\begin{table}
\centering
\caption{Comparison of real-world transactions by Liftago to SIDMAF with \(p_T^3=0.9\), probability of at least three driver accepts}
\begin{tabular}{l|cc}

   Market formation & KPI1 & KPI2 \\
  \hline
  Liftago & 0.476  & 7.67 \\
  SIDMAF & \textbf{0.828} & \textbf{6.19}\\

\end{tabular}\label{table:results_2}
\end{table}

\section{Conclusions and Future Directions}
In our simulations, we found that the data driven approach to market formation in on-demand transport is a viable alternative to heuristics based on human expertise. Our proposed algorithm, SIDMAF, outperformed heuristics used by Liftago in both observed performance indicators. These promising results suggest that SIDMAF or its derivative could be a good candidate for real world verification based on A/B testing.

Additionally, we give an overview of the features that affect the decisions of the drivers when considering whether to bid on ride order requests. While the distance to the customer and the requested ride distance are important, driver's past behaviour and preferences play similarly important role. 

\subsection*{Acknowledgements}
This work was funded by Technology Agency of the Czech Republic (grant no. TE01020155). Access to computing and storage facilities owned by parties and projects contributing to the National Grid Infrastructure MetaCentrum, provided under the programme "Projects of Large Research, Development, and Innovations Infrastructures" (CESNET LM2015042), is greatly appreciated.

\bibliographystyle{ieeetr}
\bibliography{magazine_ODT}

\begin{thebibliography}{1}

\bibitem{Breiman2001}
L.~Breiman, ``Random forests,'' {\em Machine learning}, vol.~45, no.~1,
  pp.~5--32, 2001.

\bibitem{scikit-learn}
F.~Pedregosa, G.~Varoquaux, A.~Gramfort, V.~Michel, B.~Thirion, O.~Grisel,
  M.~Blondel, P.~Prettenhofer, R.~Weiss, V.~Dubourg, J.~Vanderplas, A.~Passos,
  D.~Cournapeau, M.~Brucher, M.~Perrot, and E.~Duchesnay, ``Scikit-learn:
  Machine learning in {P}ython,'' {\em Journal of Machine Learning Research},
  vol.~12, pp.~2825--2830, 2011.

\end{thebibliography}

\end{document}